\documentclass[a4paper]{styles/svproc}
\usepackage{url}
\usepackage{mathptmx} 
\usepackage{times} 
\usepackage{amsmath} 
\usepackage{bm}
\usepackage{amssymb}  
\usepackage{graphicx}
\usepackage{mathtools}
\usepackage{upgreek}
\usepackage{gensymb}
\usepackage[noadjust]{cite}
\usepackage{placeins}
\usepackage{xcolor}
\usepackage{url}
\usepackage{mwe}
\usepackage{balance}
\usepackage{multirow}
\usepackage{caption}
\usepackage{subcaption}

\begin{document}
\mainmatter              
\title{\LARGE \bf
NASU -- Novel Actuating Screw Unit: \\
Origami-inspired Screw-based Propulsion on Mobile Ground Robots
}
\titlerunning{NASU - Novel Actuating Screw Unit}  
%

\author{Calvin Joyce\inst{*} \and Jason Lim \and
Roger Nguyen \and Michael Owens \and Sara Wickenhiser \and Elizabeth Peiros \and
Florian Richter \and Michael C. Yip}

\authorrunning{Calvin Joyce et al.} 
%
\tocauthor{Calvin Joyce, Jason Lim, Roger Nguyen,  Michael Owens, Sara Wickenhiser \\ Elizabeth Peiros, Florian Richter, Michael C. Yip}
\institute{University of California, San Diego, La Jolla CA  92093, USA,\\
\email{cajoyce@ucsd.edu}
}

\maketitle              

\begin{abstract}

Screw-based locomotion is a robust method of locomotion across a wide range of media including water, sand, and gravel.
A challenge with screws is their significant number of impactful design parameters that affect locomotion performance.
One crucial parameter is the angle of attack (also called the lead angle), which has been shown to significantly impact the performance of screw propellers in terms of traveling velocity, force produced, degree of slip, and sinkage.
As a result, the optimal design choice may vary significantly depending on application and mission objectives.
In this work, we present the Novel Actuating Screw Unit (NASU). It is the first screw-based propulsion design that enables dynamic reconfiguration of the angle of attack for optimized locomotion across multiple media and use cases.
The design is inspired by the kresling unit, a mechanism from origami robotics, and the angle of attack is adjusted with a linear actuator, while the entire unit is spun on its axis to generate propulsion.
NASU is integrated into a mobile test bed and experiments are conducted in various media including gravel, grass, and sand. Our experiment results indicate a trade-off between locomotive efficiency and velocity exists in regards to angle of attack, and the proposed design is a promising direction for reconfigurable screws by allowing control to optimize for efficiency or velocity.
\end{abstract}

\section{Introduction}

Archimedean screws were first used, in terms of locomotion mechanisms, as propellers for watercraft \cite{rossell_chapman_1962, wells_1841}, and later proposed for amphibious vehicles capable of traversing and transporting loads in both liquid and soil environments \cite{neumeyer1965marsh, fales1972riverine}. Screw-propelled vehicles and rovers have since demonstrated success across a very wide range of environments including snow, ice, sand, and other granular media \cite{Freeberg_2010, villacres2023literature}. 


These screw-based designs have great potential for exploratory robots, overcoming limitations faced by traditional wheeled rover designs and potentially avoiding situations like NASA's Spirit rover getting stuck in loose sand on Mars \cite{li_2008}.
Most screw-based designs use a parallel configuration where two counter-rotating, opposite-handed screws allow for steering \cite{Nagaoka_2010, osinski_2015}.
Quad-screw designs have been proposed to take advantage of a partial screw-slippage case for omnidirectional drive \cite{USF,lugo2017conceptual}.

\begin{figure}[t!]
    \centering
    \includegraphics[trim={0 0cm 0cm 2cm},clip, width=\linewidth]{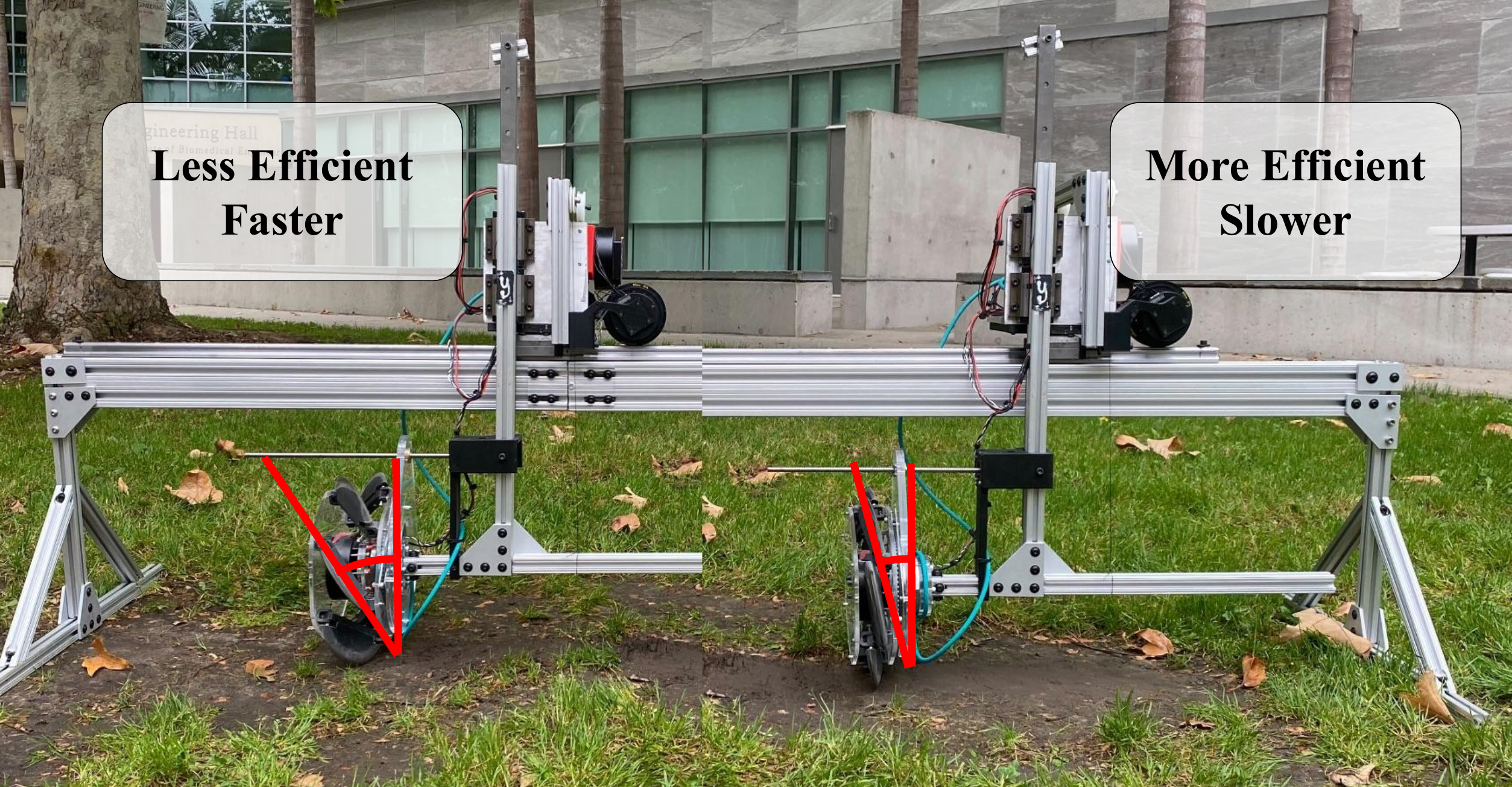}
    \caption{NASU is a screw-based propulsion unit for robot ground mobility that dynamically adjusts its angle of attack. The use of NASU is to allow control for optimizing efficiency or velocity over a wide array of terrains. It allows for trading off higher traveling velocity and lower efficiency (left) for higher efficiency and lower velocity (right) for resource-constrained robotic deployments.}
    \label{fig:system_overview}

\end{figure}

To maintain more points of contact and increase versatility, hyper-redundant snake-like designs have also been proposed. One such robot is ARCSnake~\cite{arcsnake_icra, arcsnake_tro, lim2023amphibious}, which combines Archimedean screw-based propulsion and serpentine body reshaping, and is the evolutionary precursor to the NASA Extant Exobiology Life Surveyor (EELS) robot \cite{thakker2023eels, gildner2024boldly}. EELS is anticipated to serve as a science research vehicle for both earth science missions as well as for space exploration missions on Enceladus and Europa~\cite{carpenter2021exobiology}. Deployments of the robot types mentioned above are often in resource-constrained environments where versatility and adaptability are crucial.

\subsection{Contributions}

Previous research has shown that a screw propeller's angle of attack plays a crucial role in locomotion performance, affecting traveling velocity, thrust force, degree of slip, and sinkage. The optimal design choice may depend heavily on the desired use case \cite{cole1961inquiry, dugoff1967model, Nagaoka_2010}.
In pursuit of a screw propulsion mechanism that can adapt to different use cases, we present the Novel Actuator Screw Unit (NASU), a screw propulsion design that enables an adjustable angle of attack inspired by an origami kresling unit \cite{novelino2020untethered, Miyazawa2022, ze2022soft, PhysRevE.101.063003}.
However, unlike other origami robots which are often made from compliant materials, NASU is made from rigid material to withstand the high loads encountered in screw-based locomotion \cite{icra2023}. 
Previously Ze et. al. developed an origami kresling unit for locomotion through iterative compression and decompression creating a sliding motion.
In contrast, NASU relies on traditional screw propulsion for locomotion and uses its origami nature to adjust the screw-blade's angle of attack within a range of 10-35\textdegree{}.
This on-the-fly re-configuration enables intelligent control for optimizing the angle of attack as required for different objectives (see Figure \ref{fig:system_overview}). 
Ultimately, NASU offers a novel screw-based locomotion approach capable of dynamically adjusting the angle of attack for various multi-domain applications.

\subsection{Related Works}

A challenge for screw-based locomotion designs is the significant number of design parameters that can affect performance.
Exploration has been done into the effect of varying parameters such as angle of attack (also referred to as lead angle, pitch angle, and helix angle in previous literature), total screw length, blade height, and number of starts. 

Some of the first studies in screw propulsion and locomotion showed theoretically and experimentally that velocity tends to increase with increasing angle of attack, while thrust force decreases in sand and soil-like environments \cite{cole1961inquiry, dugoff1967model}. Later computational and experimental analyses confirmed a higher thrust force is produced with smaller angles of attack in the case of granular media composed of silica beads \cite{marvi_2018, marvi_2019}. In contrast, higher angles of attack tend to increase the amount of thrust force in fluid environments \cite{cole1961inquiry, Mayfield_2015, lim2023amphibious}. A longer total screw length also leads to increased performance in water \cite{cole1961inquiry}.  
An in-depth study into screw-terrain interaction using terramechanics models determined that the ratio of blade height to screw radius is a vital factor for performance in soft granular media as it affects the level of sinkage and contact surface area. That angle of attack should be chosen carefully to ensure that the screws can produce a positive tractive force when slipping \cite{Nagaoka_2010}. In an analysis of screw performance in the sand across different parameter combinations for the angle of attack, blade height, and number of starts, optimal values were found to be 35\degree, 14 mm, and 1, respectively \cite{seo_2021}.


Outside of screw-based locomotion, reconfiguration for aerial locomotion has been explored in nature to understand how changing parameters of bird wings affect flight performance \cite{doi:10.1177/1045389X11414084, doi:10.1098/rsif.2017.0240}.
Similar abilities have been replicated in robotics to achieve more efficient flight performance from wing designs \cite{Wichita}.
Reconfigurable propeller designs have also been proposed to create more efficient propulsion in different scenarios \cite{10.1115/1.4054249, CHEN2017746}.
Adjustable diameter wheels have been proposed to be able to climb over different heights of obstacles while still being able to get into small places for search and rescue applications \cite{nagatani2007development}.
Similarly, NASU can dynamically change the angle of attack to adjust for different locomotion capabilities.


\section{System Design}

\begin{figure}[!t]
    \centering
    \includegraphics[width=0.49\linewidth, trim={0 0 0 0},clip]
    {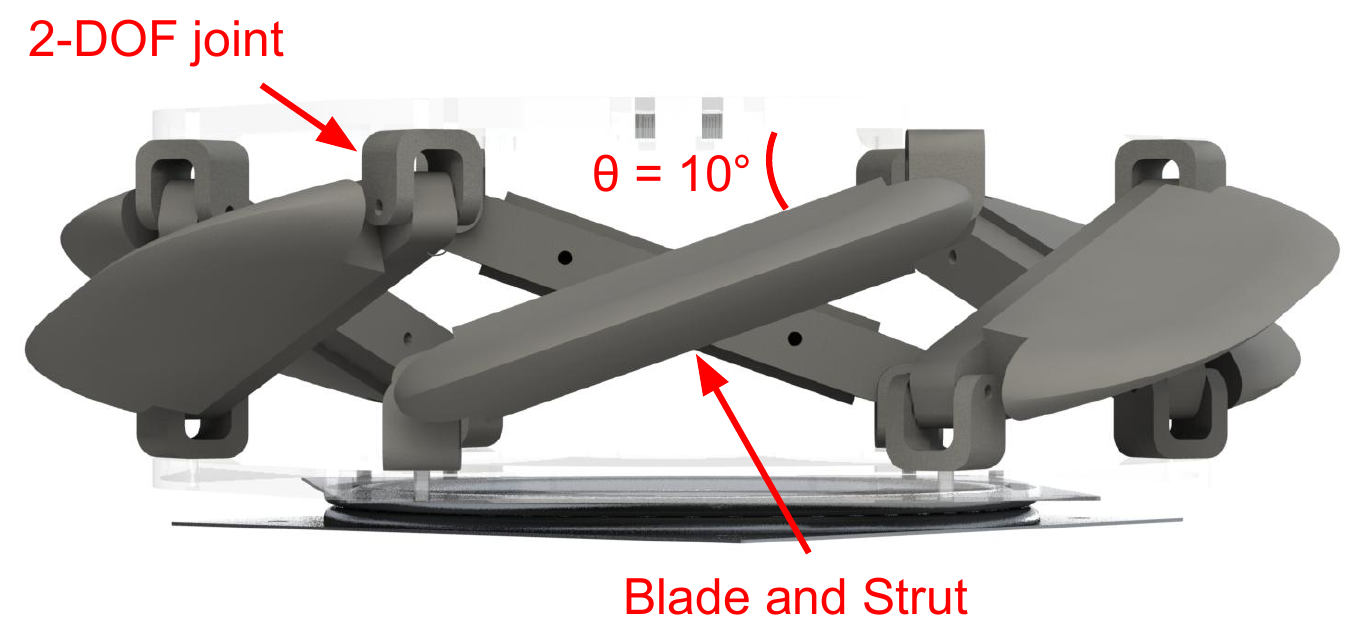} 
    \includegraphics[width=0.49\linewidth, trim={0 0 0 0},clip]
    {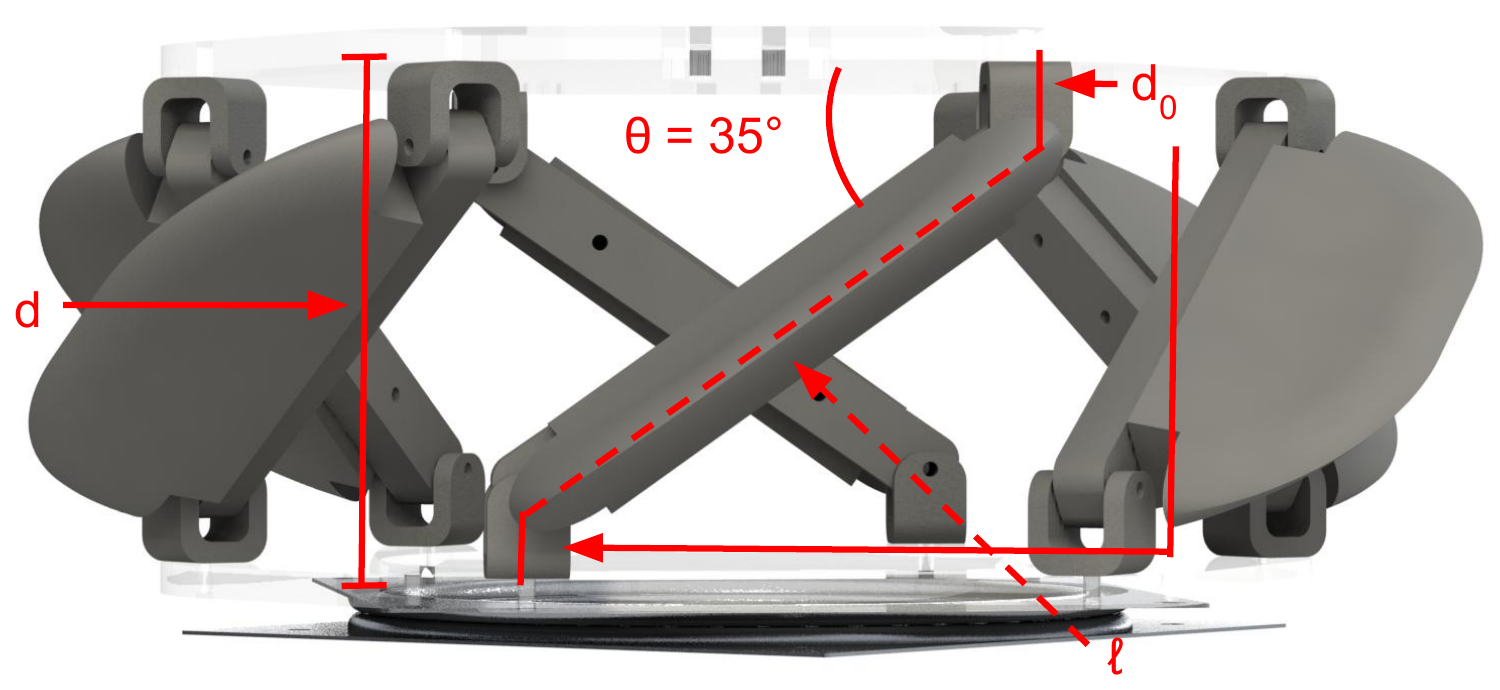}
    \caption{From left to right, the figures show NASU at the minimum and maximum angles of attack.
    The angle of attack is set by linearly actuating the back plate whose position is denoted with parameter $d$.
    When the linear actuator moves, 2-DOF joints connecting the front and back plates to the struts are passively actuated to set the angle of attack of the blades.
    Finally, the angle of attack can be computed using $d_0$ and the strut length, denoted as $\ell$.
    }
    \label{fig:nasu_angles}
\end{figure}

The design of NASU, as noted previously, is structured after the origami kresling unit.
The kresling unit defines a kinematic constraint between the length of the unit and the angle of the struts. By augmenting and controlling the total length of the unit, we can control the angle of attack. Its operation is shown in Figure \ref{fig:nasu_angles}.
With our design, we verified the ability to produce appropriate thrust force and withstand reaction forces based on previous literature that captured data on screw forces. 
Finally, we integrated the NASU into a mobile testing platform which provided a means to measure performance across media.

\subsection{Mechanical Design}

A linear actuator drives the position of the back plate which sets the length of the unit and thus the angle of attack as shown in Figure \ref{fig:nasu_angles}.
When the linear actuator moves to set the distance between the plates, 2-DOF joints connecting the front and back plates to the struts are passively actuated to set the angle of attack on the blades.
This process is described in the following equations:
\begin{align}
    d &= d_0 + \ell\sin\left(\theta\right) \\
    \theta &= \sin^{-1}\left(\frac{d - d_0}{\ell}\right)
\end{align}
where the distance between the inner surfaces of the plates is denoted by $d$, $d_0$ is the sum of the offsets between the plates and the pivot points, $\ell$ is the length of the strut, and $\theta$ is the angle of attack.

Our design is set to have an adjustable angle of attack, $\theta$, with a range of 10-35\textdegree{}, and strut length of $\ell = 100 \mathrm{mm}$. These parameters were chosen to match similar ratios with previous literature \cite{arcsnake_tro, USF}.
This results in a linear actuation range, $d$, of 48-88m given our 2-DOF design with $d_0=31 \mathrm{mm}$. The blade design in terms of height, shape, and profile was chosen based on previous work \cite{crawling}.
Furthermore, the design constraints set the root radius, $r$, of the NASU to 192mm.
Finally, we choose a 6-blade design to minimize the discontinuities of contact with the environment which is discussed in more detail in the next subsection.

\begin{figure}[t]
    \centering
    \includegraphics[width=0.49\linewidth]{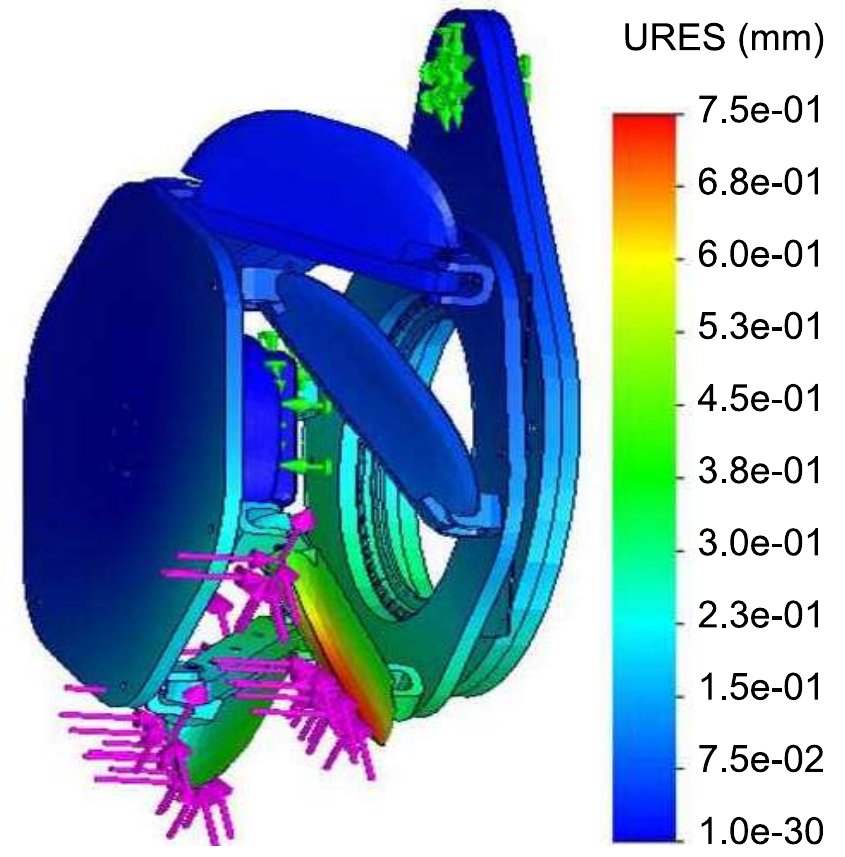}  \includegraphics[width=0.49\linewidth]{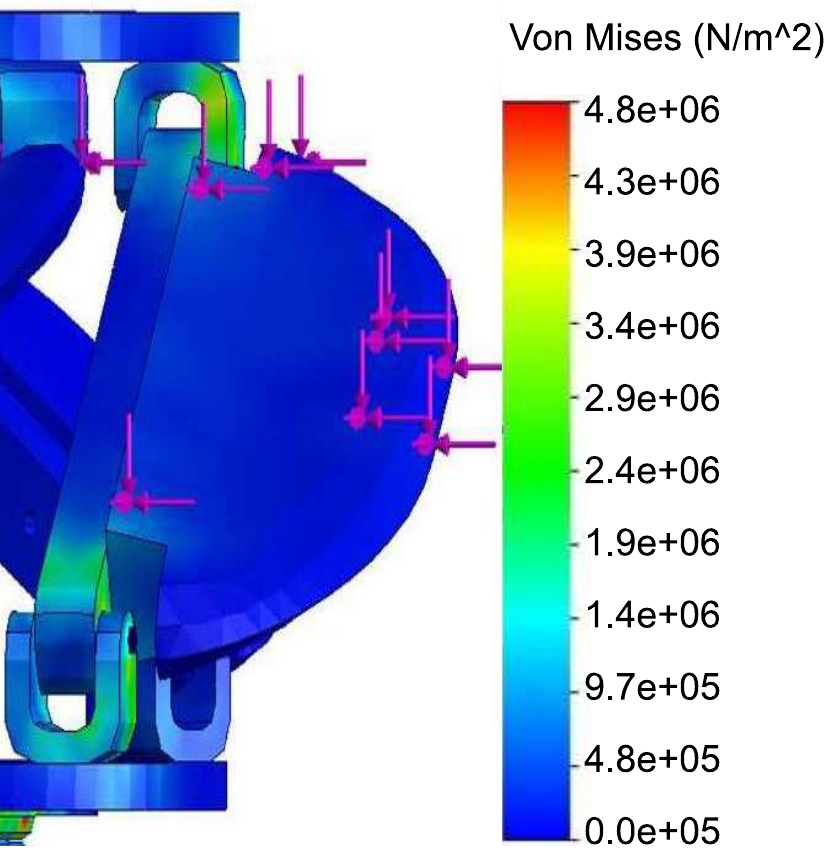}
    \caption{
    From left to right, the figures show an FEA estimating the displacement of the entire NASU and stress on one of the blades and its 2-DOF joints, respectively.
    The applied force is taken from the maximum amount of force measured in our previous outdoor screw experiments \cite{icra2023}.
    Fixture points are set as connections from NASU to the ball screw which drives NASU's ability to change the angle of attacks.
    NASU was set to its maximum angle, 35\textdegree{}, to simulate the most extreme possible situation and resulted in minimal deflection and stress.}
    \label{fig:NASU_fea}
\end{figure}

Manufacturing of NASU was done with a combination of 3D printing from an Onyx One printer with the Onyx material, a carbon fiber-filled thermoplastic, and acrylic parts that were laser cut.
We also added replaceable blade attachment points along the struts so that in the future we can alter the blade height, profile shape, and length.
The linear actuator used to set the distance between the front and black plates, $d$, is a linear ball screw and a Nema-17 stepper motor. 



Finite Element Analysis (FEA) was conducted to ensure the integrity of our design using a static Solidworks Simulation model.
The maximum force the blades will encounter is a static hold at the peak force before the media shears.
In the simulation, an angle of 35\textdegree{} was chosen as it corresponds to the largest resultant reaction force.
The force was applied across the blade tip and surface while the fixed point was set as the connection between NASU and the ball screw.
Pin connections were used to model the correct degrees of freedom of the 2-DOF joint.
The external load applied is taken from our previous work where we recorded a maximum force of $[4.71 N,\;20.74N,\;8.41N]$ when testing in various media \cite{icra2023}.
Note that +Z is in the propulsion direction of NASU and +X is aligned with gravity. 
In the previous works' experiments, gravity was removed by taring the senor, thus we added a representative gravity back into the X vector by using the mass of our NASU design, 2.1kg.
The estimated applied forces for the simulation are $[15.5 N,\;20.74N,\;8.41N]$.
The results of the FAE are shown in Figure \ref{fig:NASU_fea} and 
a minimum Factor of Safety (FOS) of 9.9, a max displacement of 0.75 mm, a max stress of $4.825 \times 10^{6}$ N/m$^2$, and a maximum strain of $4.58 \times 10^{-5}$ ESTRN were recorded.


\subsection{NASU Locomotion Modeling and Metrics}

\begin{figure}[t]
    \centering
    \includegraphics[angle=270, width=0.99\linewidth] {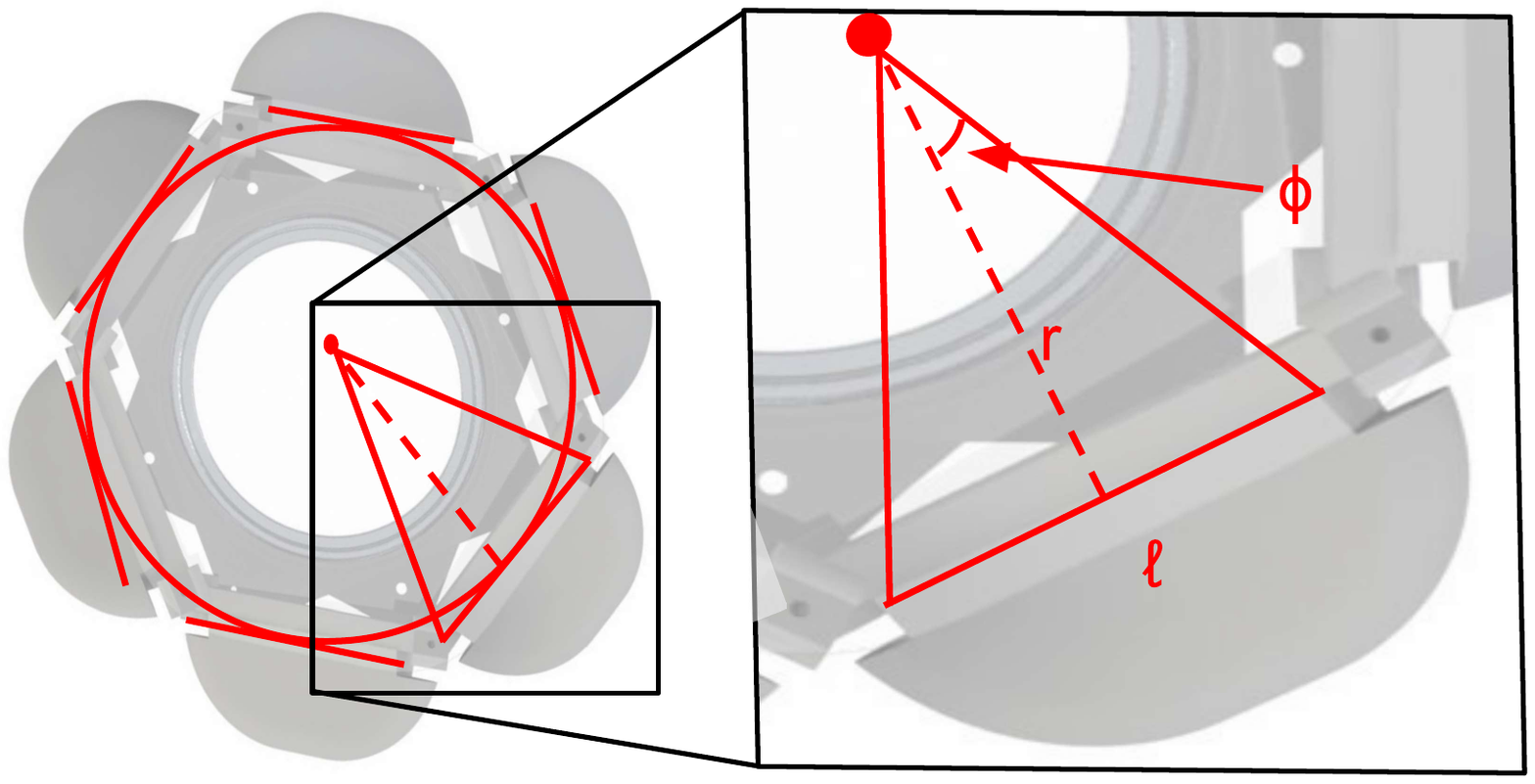}
    \caption{
    The figure shows a geometrical breakdown of a top-down view of NASU to find the percentage of blade contact with the environment for an ideal locomotion velocity.
    The circle on the left figure defines the root radius of NASU, and the struts projected on this top-down view are tangent to this circle.
    A right triangle is formed from this projection to compute the interior angle, $\phi$, which is used to solve for the percentage of blade contact.
    }
    \label{fig: Non_ideal_calc}
\end{figure}

The ideal locomotion velocity for a screw assumes the screw threads move through grooves as if they were in a threaded nut.
Since NASU is designed with a series of short blades rather than a continuous helix, we relax the continuous contact assumption to derive the following ideal velocity equation:
\begin{equation}
    v_{ideal} = r K_c(\theta) \omega \tan(\theta)
    \label{eqn: ideal}
\end{equation}
where $r$ is the inner radius of the screw, $\omega$ is the angular velocity of the screw motor, and $K_c(\theta)$ represents the percent of contact with the media it is propelling in.

To calculate $K_c(\theta)$, we project the length of each blade onto the front plane of the NASU unit.
This is visually represented in Figure \ref{fig: Non_ideal_calc}, and a geometric calculation is done to find the percentage of the outer radius that would be considered in contact with the ground relative to its ideal counterpart. The simplified projection equation is shown below:
\begin{equation}
    \ell_p = \ell \cos (\theta)
    \label{eqn: projection}
\end{equation}
where $\ell_{p}$ is the projected length, $\ell$ is the blade length, and $\theta$ is the angle of attack. Next, we need to calculate the percentage of contact shown below:
\begin{align}
    &\phi = \tan^{-1}\left(\frac{0.5\ell_{p}}{r} \right) \\
    &K_c(\theta) = \frac{n \phi}{\pi}
    \label{eqn: percentage_ideal}
\end{align}
where $\phi$ is one of the angles formed by our right-angle triangles depicted in Figure \ref{fig: Non_ideal_calc}, $r$ is the inner radius of the circle defined by the pivot points of the 2-DOF joints (i.e. root radius), and $n$ is the number of blades.

The metrics that we use to characterize performance are linear traveling velocity and locomotive efficiency.
Locomotive efficiency is defined as follows:
\begin{equation}
    \eta_m = \frac{ F_{thrust} v}{\tau_{in} \, \omega}
    \label{eqn: efficiency}
\end{equation}
where $F_{thrust}$ is the force produced by the NASU along its longitudinal axis, $v$ is the linear traveling velocity, and $\tau_{in}$ and $\omega$ are the average input torque and angular velocity, respectively.


\begin{figure}[!t]
  \centering
  \begin{minipage}{\textwidth}
    \centering
    \subcaptionbox{Full Attachment of NASU on Mobile Test Bed \cite{icra2023}}
    {
      \includegraphics[width=.8\textwidth]{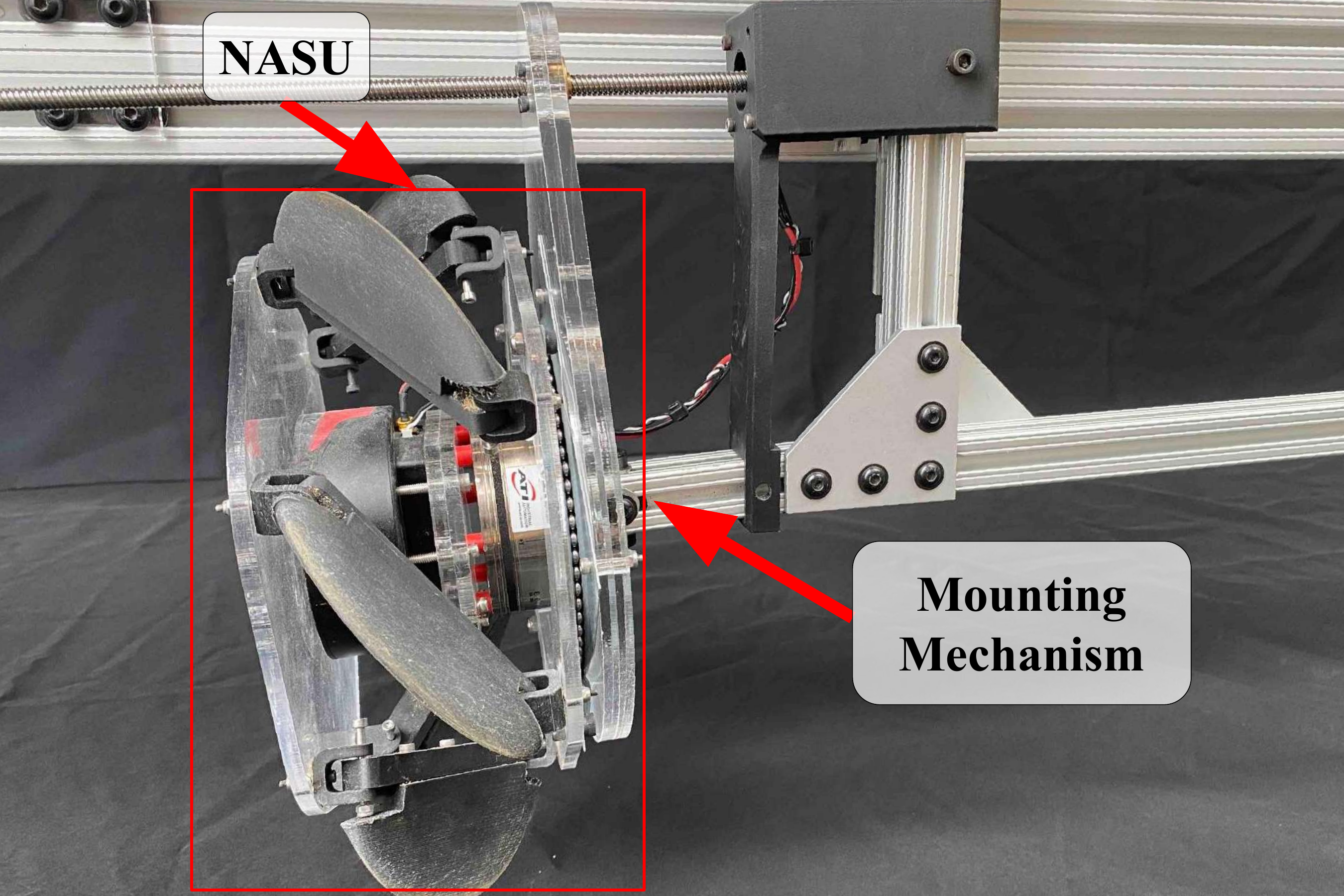}
      \label{fig:test bed_mod_a}
    }
    \vspace{1mm}
  \end{minipage}
  \begin{minipage}{.8\textwidth}
    \centering
    \begin{minipage}{0.25\textwidth}
      \subcaptionbox{Pulley System}
      {
        \includegraphics[width=\textwidth]{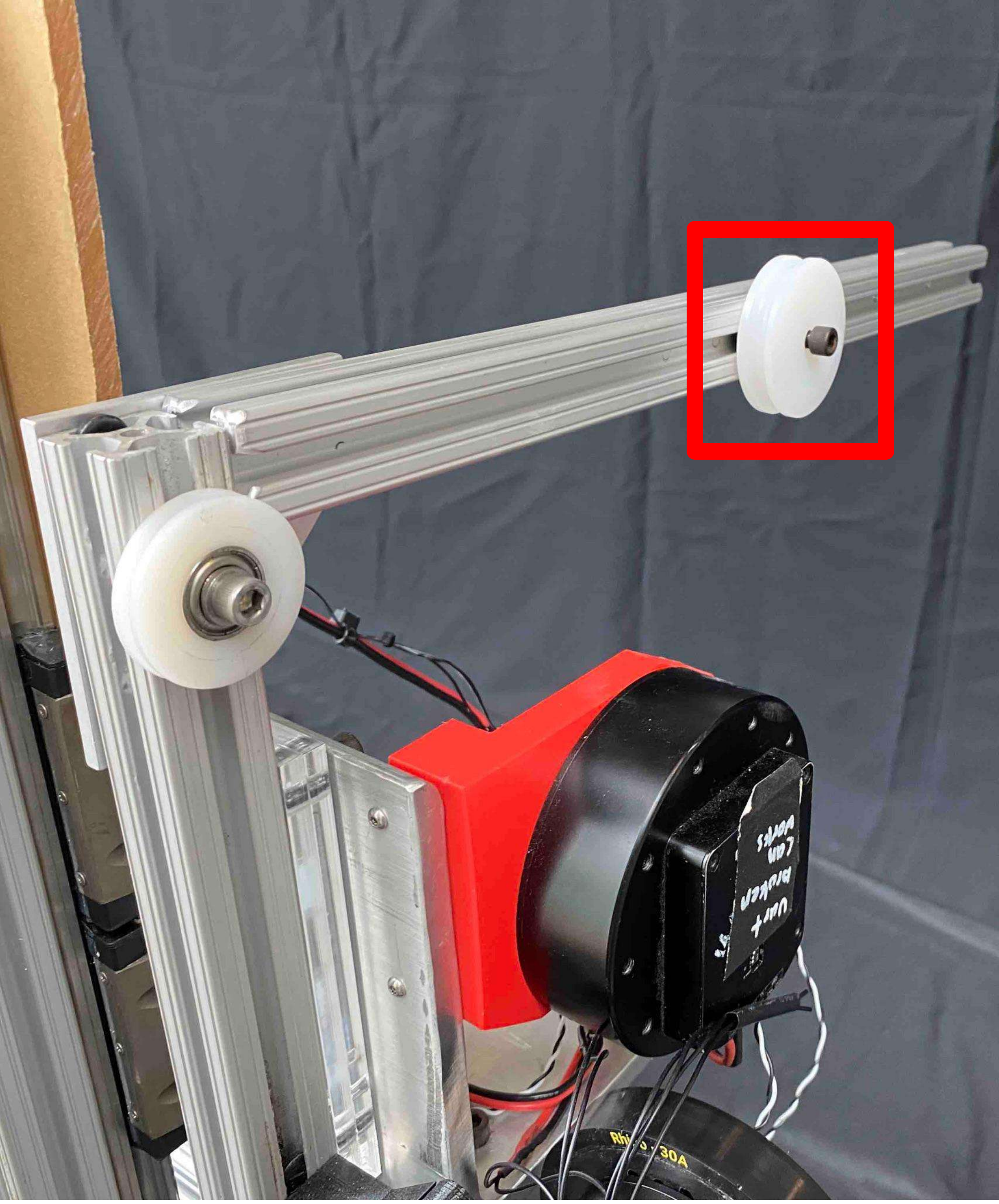}
        \label{fig:test bed_mod_b}
      }
    \end{minipage}
    \hfill
    \begin{minipage}{0.25\textwidth}
      \subcaptionbox{Linear Actuator}
      {
        \includegraphics[width=\textwidth, trim={0 0 0 4.5cm}, clip]{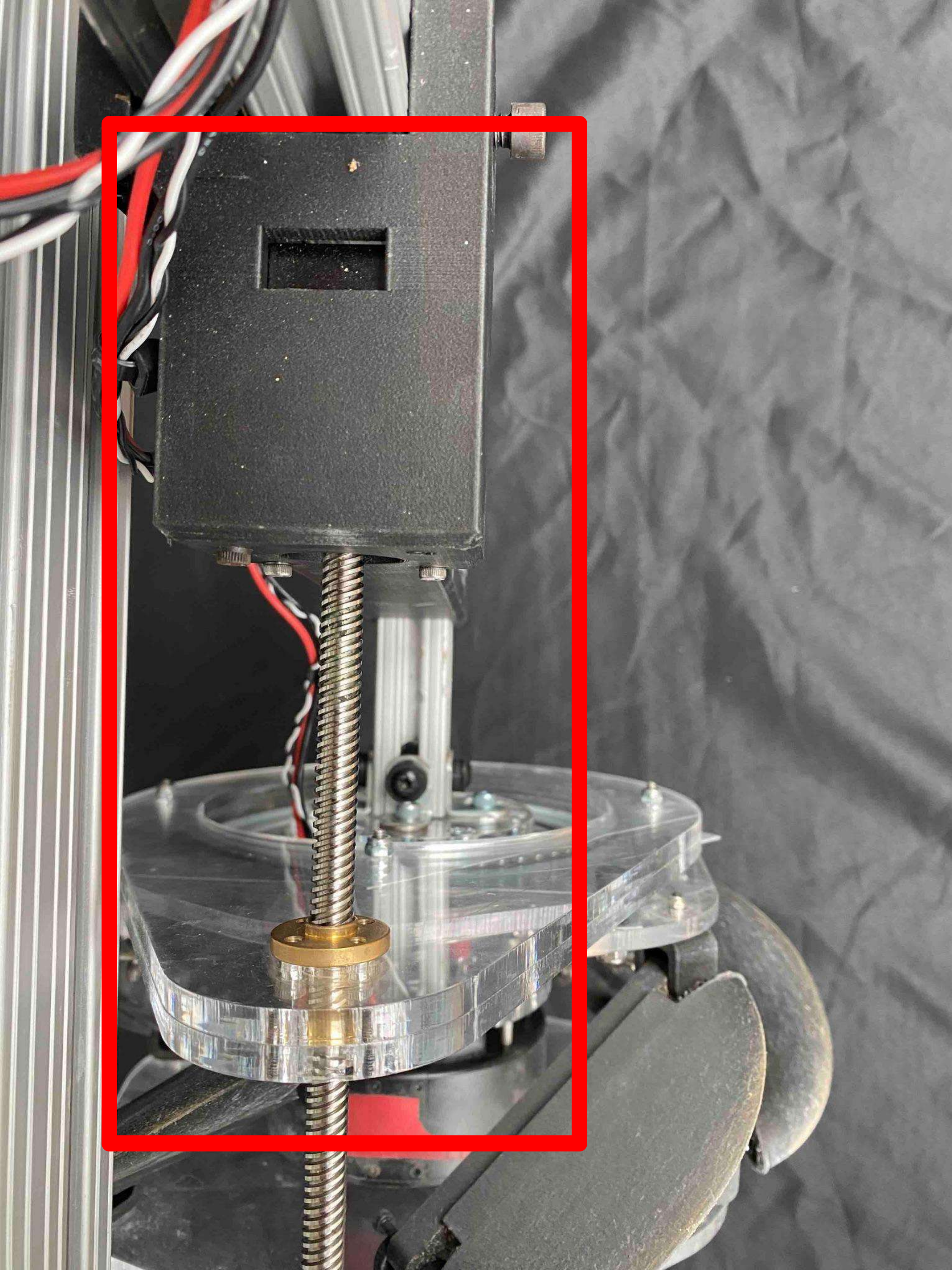}
        \label{fig:test bed_mod_c}
      }
    \end{minipage}
    \hfill
    \begin{minipage}{0.25\textwidth}
      \subcaptionbox{Bearing}
      {
        \includegraphics[width=\textwidth, trim={0 2cm 0 0}, clip]{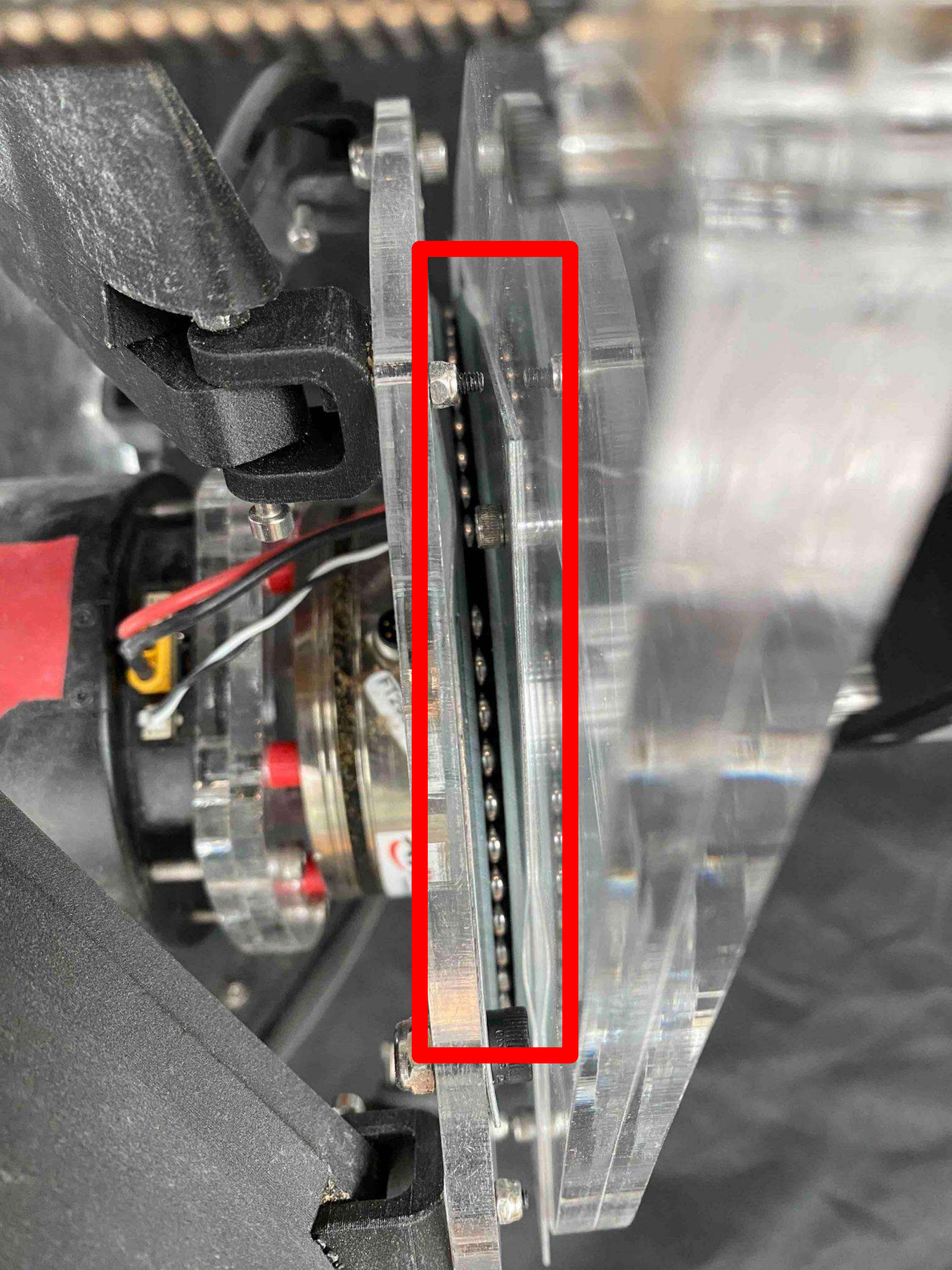}
        \label{fig:test bed_mod_d}
      }
    \end{minipage}
  \end{minipage}
  \caption{
    The mobile test bed was augmented to mount NASU for experimentation.
    The mobile test bed constrains the motion linearly and measures the resultant propulsion forces and traveling velocity.
    The pulley system is used to apply a counterweight compensating for the test bed's added mass which leaves the effective mass equal to that of NASU including its driving motor and linear actuator to adjust the angle of attack.
  }
  \label{fig:test bed_mod}
\end{figure}

\subsection{Mobile Test Bed: NASU Actuation}

We reuse the mobile test bed from our previous work \cite{icra2023} with an attachment for NASU  for our multi-media experiments.
Through the mobile test bed, NASU is constrained to travel in a single, linear direction by linear rails.
Traveling velocity is measured on the linear-travel rail with a passive RMD-L 7015 motor.
Meanwhile, the height-adjusting linear rail helps compensate for potentially uneven surfaces being experimented on.
Finally, a 6-DOF FTS, Axia80 (ATI Industrial Automation), is positioned near the center of mass to measure the screw's applied torque and resulting screw-locomotion forces.

Modifications were made to the test bed to enable proper testing of NASU, and Figure \ref{fig:test bed_mod} shows an overview of our changes to the test bed.
First, a pulley system is implemented on the test bed to apply a counterweight that compensates for the additional mass not part of the NASU design (e.g. vertical bar, linear rail, 6-DOF FTS).
The NASU unit by itself weighs 2.1kg; meanwhile, the entire mounted NASU system on the test bed weighs 6.2kg.
Therefore, we apply 4.1kg of counterweight on the pulley, and this feature can be seen in Figure \ref{fig:test bed_mod}(b).
Second, the motor housing for the linear actuator is secured to the vertical bar of the test bed as seen in Figure \ref{fig:test bed_mod}(c).
A constraining plate is placed behind the back plate of NASU with its orientation constrained but linearly motion allowed.
The ball screw drives the shortening or elongation of the mechanism through the movement of this plate.
Finally, for the kresling unit design to properly morph to new angles of attack it was necessary to maintain a torsional degree of freedom which was accomplished through the addition of a lazy susan bearing sandwiched between the back plate of the NASU unit and the constraining plate driven by the linear actuator.
The bearing is shown in Figure \ref{fig:test bed_mod}(d).

\begin{figure*}[!ht]
    \centering
    \includegraphics[width=0.135\linewidth]{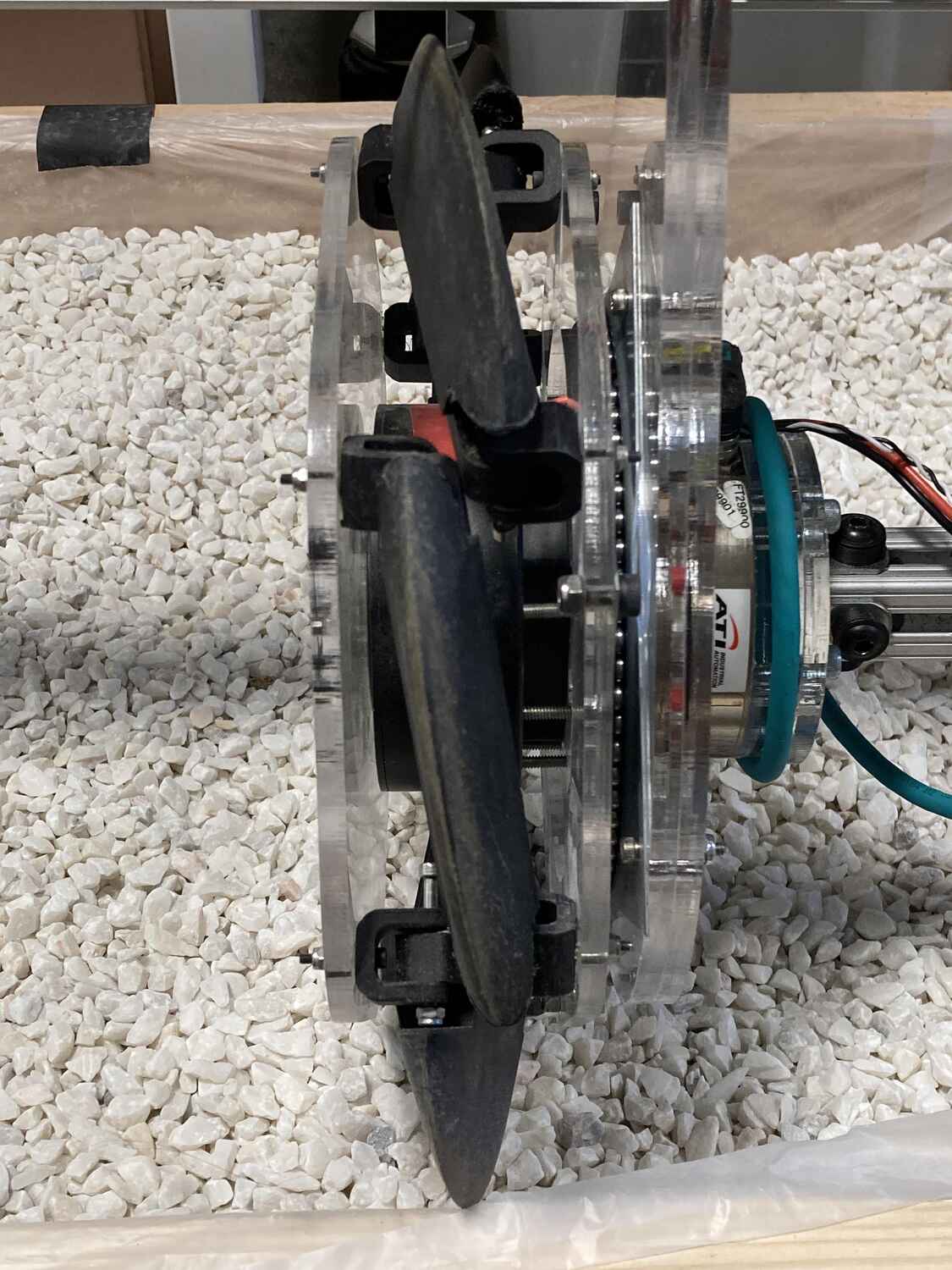}
    \includegraphics[width=0.135\linewidth]{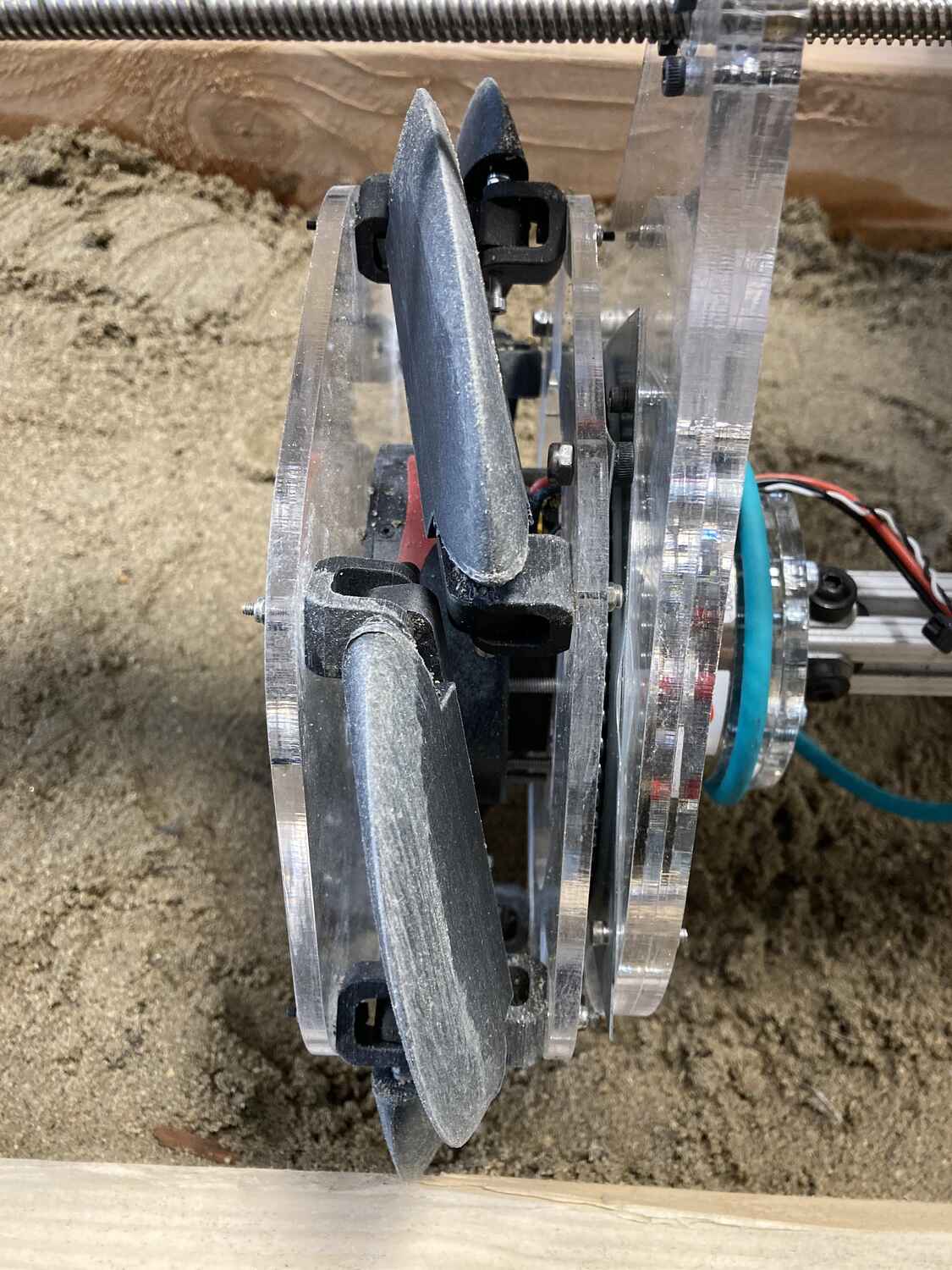}
    \includegraphics[width=0.135\linewidth]{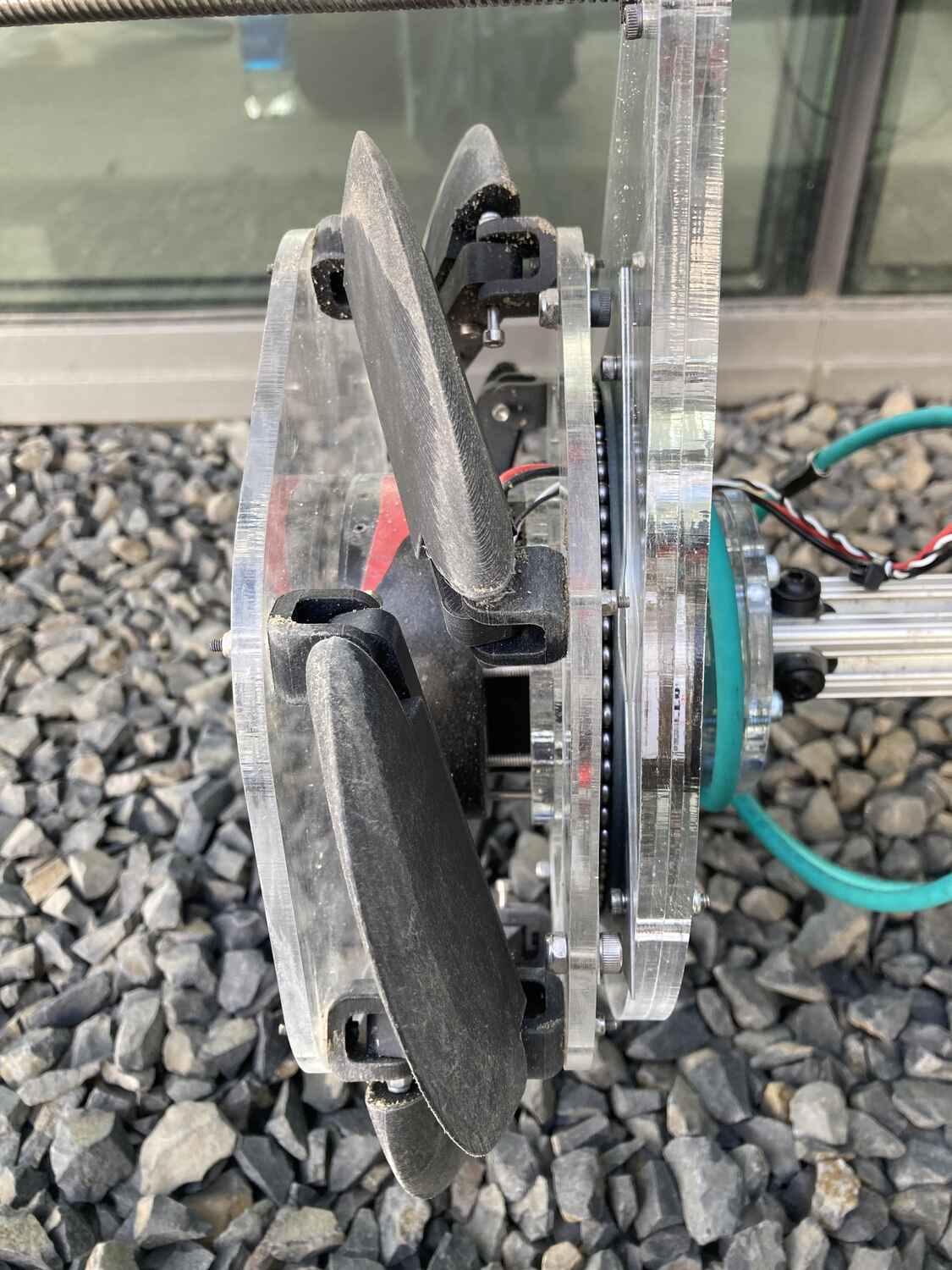}
    \includegraphics[width=0.135\linewidth]{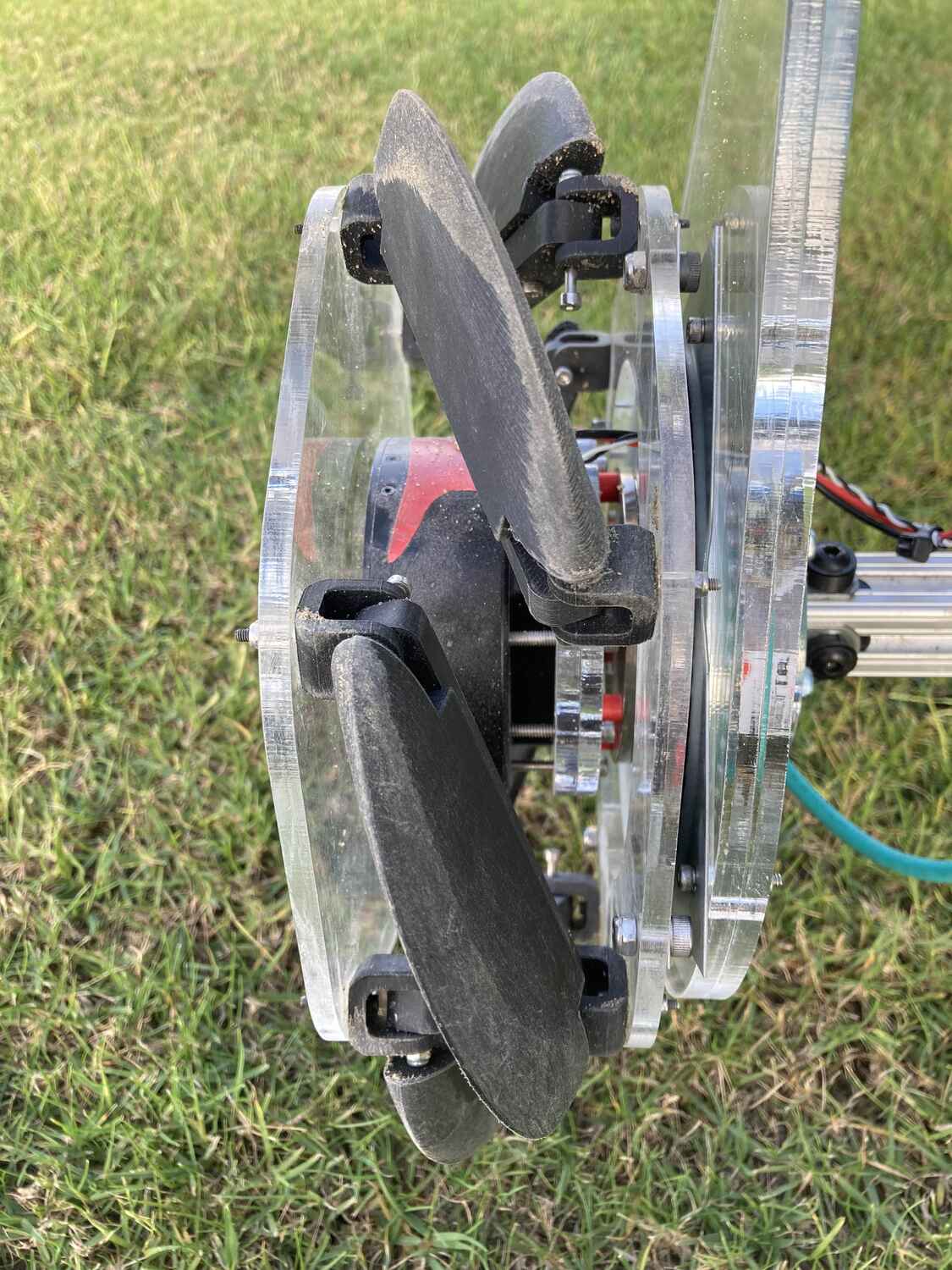}
    \includegraphics[width=0.135\linewidth]{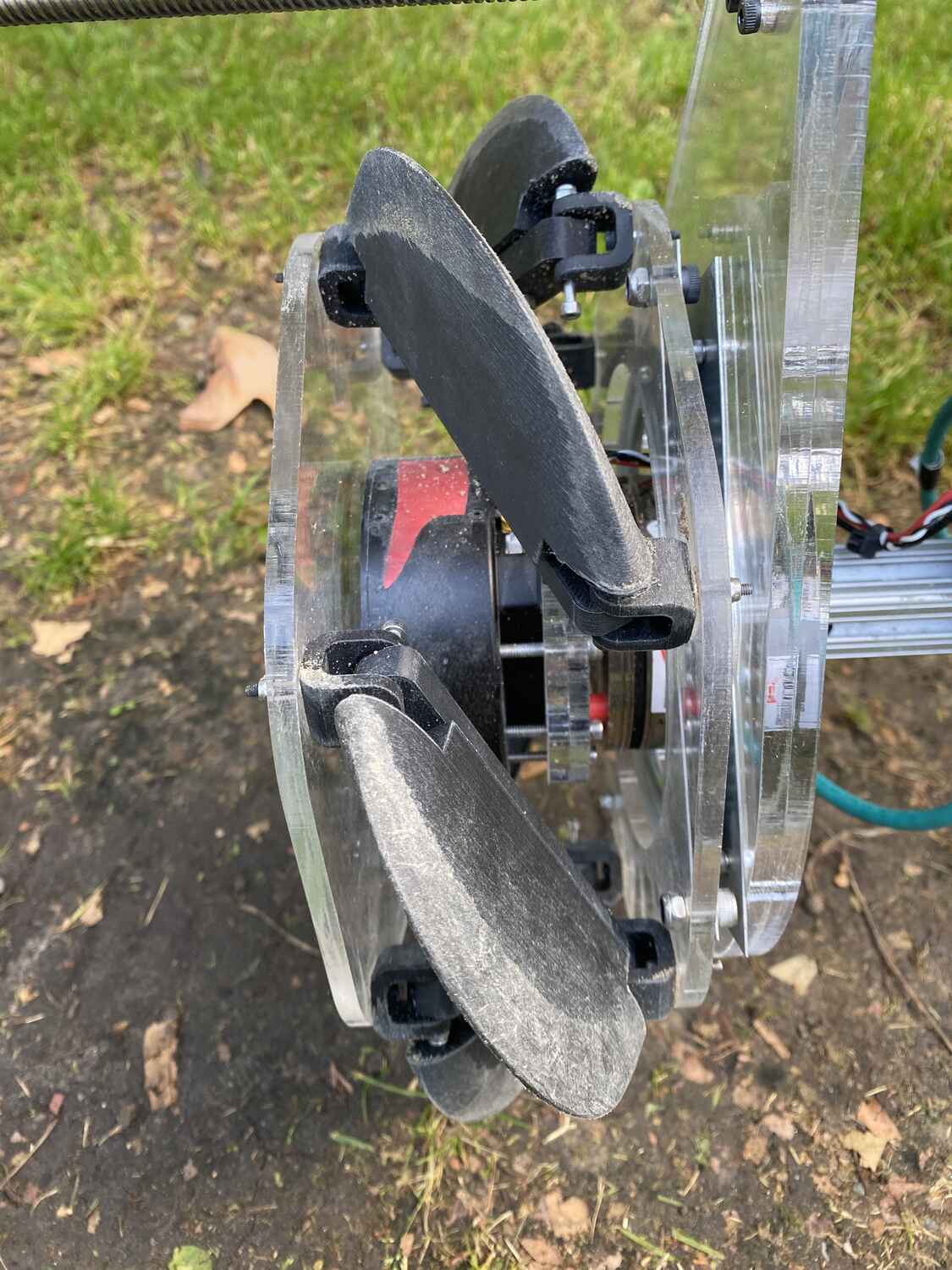}
    \includegraphics[width=0.135\linewidth]{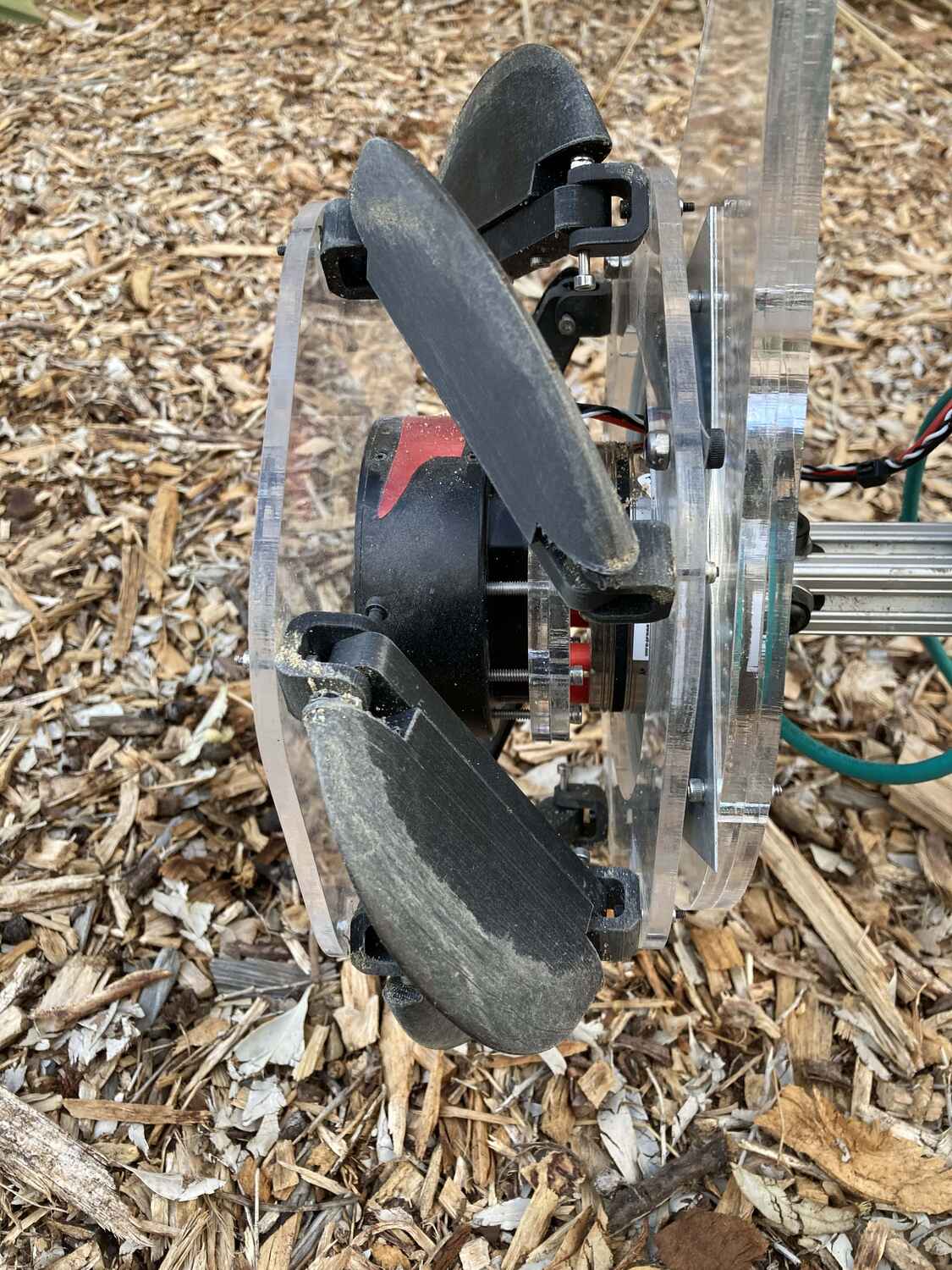}
    \includegraphics[width=0.135\linewidth]{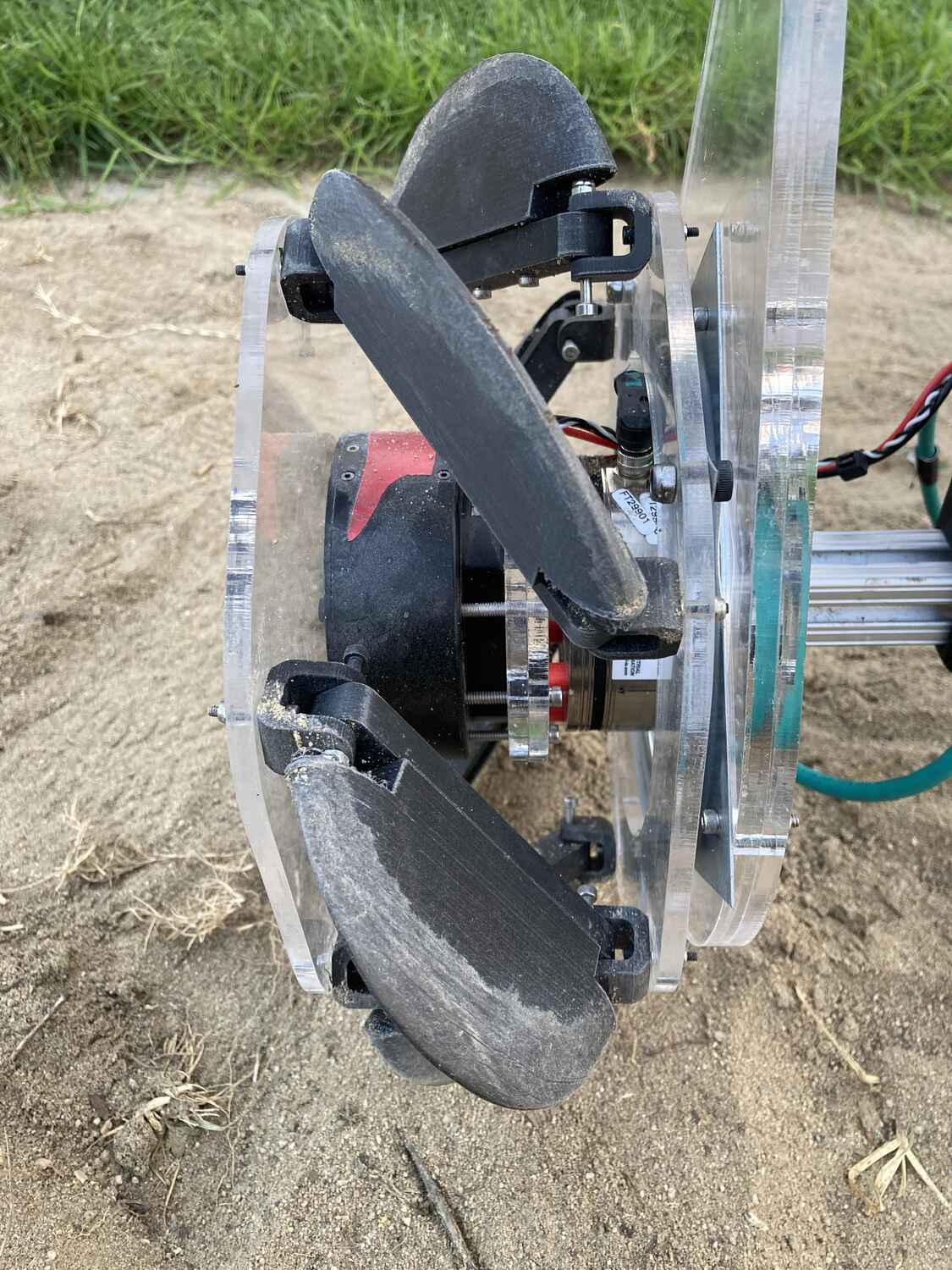}
    \caption{Using our previously developed mobile test bed that evaluates screw-based locomotion \cite{icra2023}, we experiment over various media to validate NASU's performance.
    The images from left to right are from the experiments conducted in the following media: small gravel, wet sand, big gravel, grass, mud, wood chips, and sand.
    }
    \label{fig:collage_from_experiments}
    \vspace{2mm}
    \centering
    {\includegraphics[width=0.49\linewidth]{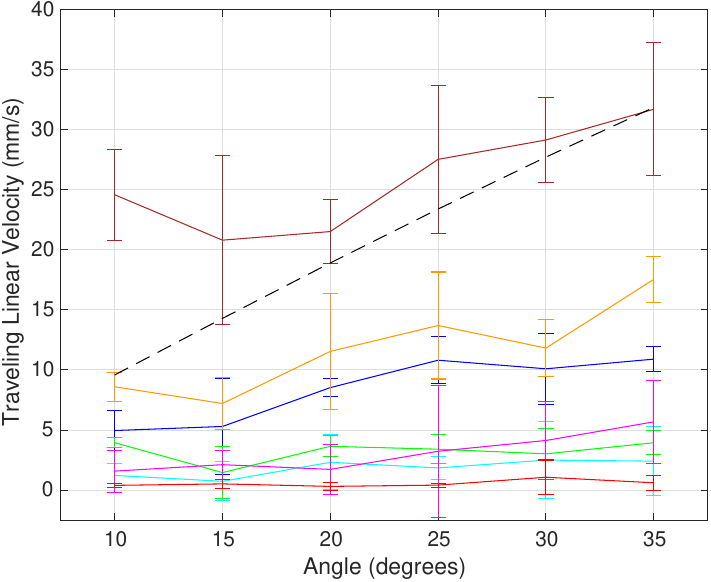}}
    \includegraphics[width=0.49\linewidth]{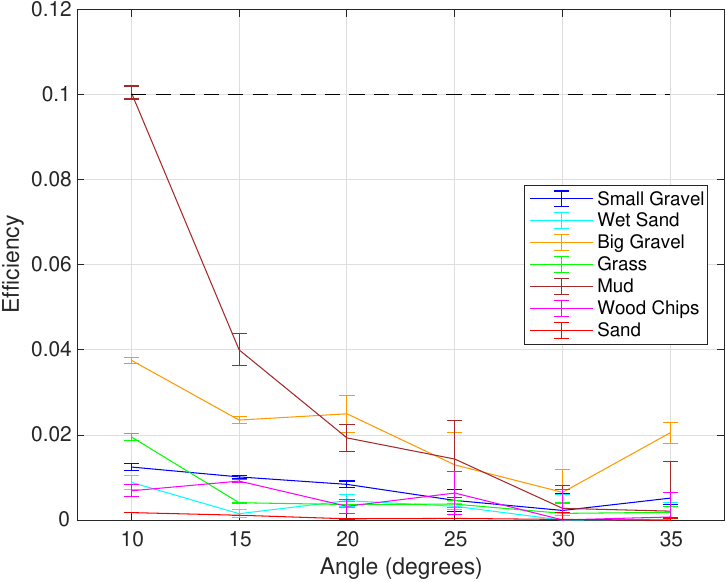}
    \caption{The left and right plots show the traveling linear velocity and efficiency, respectively, of NASU for different angles of attack set by the origami-inspired NASU system in various media.
    The dashed lines represent 25\% of the ideal velocity and 10\% efficiency in the velocity and efficiency plots, respectively.
    The trade-off between traveling velocity and efficiency as the angle of attack changes can be seen in our results and is most prominent in mud and gravel.
    }
    \label{fig:results}
\end{figure*}

\section{Experiments and Results}
\label{exp_and_res}

Experiments are conducted with the NASU on a mobile test bed in a wide range of media in both a lab setting and in real-world environments as shown in Figure \ref{fig:collage_from_experiments}.
These experiments characterize the locomotion performance at varying angles of attack on each media to provide comprehensive efficiency results of NASU.

\subsection{Experimental Setup}

The experiment is conducted on our previously developed mobile test bed \cite{icra2023} and a similar experimental setup procedure is employed:
\begin{enumerate}
    \item Flatten and level the media to achieve as uniform conditions as possible.
    \item The height-adjustable linear rail is locked such that the NASU is free-hanging. A ``free hanging" measurement from the FTS is taken to capture any potential drifts between trials. The angle of attack is set to the desired value.
    \item The height-adjustable linear rail is unlocked and the NASU is set down on the media. A ``set down" measurement from the FTS is taken to measure any pre-loading from the media on the NASU.
    \item The FTS sensor is zeroed to measure differential measurements. 
\end{enumerate}
From this point, the screw motor can be driven to begin the experiment.

To process the raw data from the test bed for analysis, the data from the motors and FTS are passed through a low pass filter with a cutoff frequency of 5 Hz and a sampling frequency of 125 Hz.
Note that the cutoff frequency is significantly smaller than the sampling rate hence the post-processing will be an effective noise reduction technique without distorting the signal.
Even after the filtering, there is substantial noise in our measurements due to conducting experiments in real, outdoor environments.
Therefore, we use the average velocity over the entire trial and maximum force within the trial to compute the highest achievable efficiency our novel mechanism can produce on that media. 
The data is also clipped manually to only include the steady-state portion of the experiment.
The test bed allows freedom of motion in both the axial and vertical directions so that the speed of travel can be measured and sinkage into the media is still allowed. In contrast, all other degrees of freedom are restricted by the test bed.
The angles of attack being tested are 10\textdegree{}, 15\textdegree{}, 20\textdegree{}, 25\textdegree{}, 30\textdegree{}, and 35\textdegree{}, and the angular velocity of NASU is set to 3.33 rad/s.

\subsection{Results}


The results of the traveling velocity and efficiency are shown in Figure \ref{fig:results}.
As is expected from theory and previous research \cite{icra2023, cole1961inquiry, dugoff1967model}, there is a positive correlation between the angle of attack and forward velocity.
This is a natural result of screw geometry and conservation of energy principles; in solid media, a higher angle of attack corresponds to a longer distance traveled per revolution.
Interestingly, our results demonstrate a negative correlation between the angle of attack and locomotive efficiency.
Thus a key finding of our results is that, in regards to varying the angle of attack, there exists a trade-off between speed and efficiency that is consistent across all of the media we tested.

\section{Discussion}

As seen in the experimental results, NASU's operating range for angle of attack, 10-35$^{\circ}$, provides a trade-off between locomotion velocity and efficiency.
While we anticipated a positive correlation between the angle of attack and forward velocity based on previous research in screw-based locomotion \cite{icra2023, cole1961inquiry, dugoff1967model}, the negative correlation with efficiency was a finding from our experimental results.
We believe the efficiency trend can be explained by several factors:
\begin{enumerate}
    \item As the angle of attack increases, the relative motion of the material increases meaning the material is moved backward faster, driving the increase in velocity. However, the blades push more material to the sides, excavating rather than contributing to thrust, thus losing far more energy to the displacement of media.
    \item For a given input torque, the available thrust force increases with decreasing angle of attack compensating for the decrease in speed. This means the towing capacity of the robot increases at the cost of moving quickly.
\end{enumerate}
Figure \ref{fig:system_overview} gives a simple diagram comparing the maximum and minimum angles of attack, as well as labeling the trade-off.
This trade-off reinforces the importance of the angle of attack in the performance of screw-based locomotion and suggests that the angle of attack should be varied depending on the desired use case, highlighting the future utility of NASU.

We find that the relative performance across different media types compares well with our previous research on screw-based locomotion performance \cite{icra2023}, which suggested that shearing force and coefficient of friction are two main properties contributing to variance in performance across media.
In this work, the best performance was obtained in mud, which was compact enough to provide a high shearing force, while still allowing the blades to sink in smoothly and gain traction well. Big gravel and small gravel exhibit the next best performance, respectively.
These media also provide a high shearing force and have a lower coefficient of friction compared to other media, although the granularity of the gravel leads to more sporadic, bumpy movement and energy lost to vertical instead of horizontal motion.

\begin{figure}[t]
    \centering
    \includegraphics[width=0.48\linewidth, trim={5cm 2cm 0 2cm}, clip]{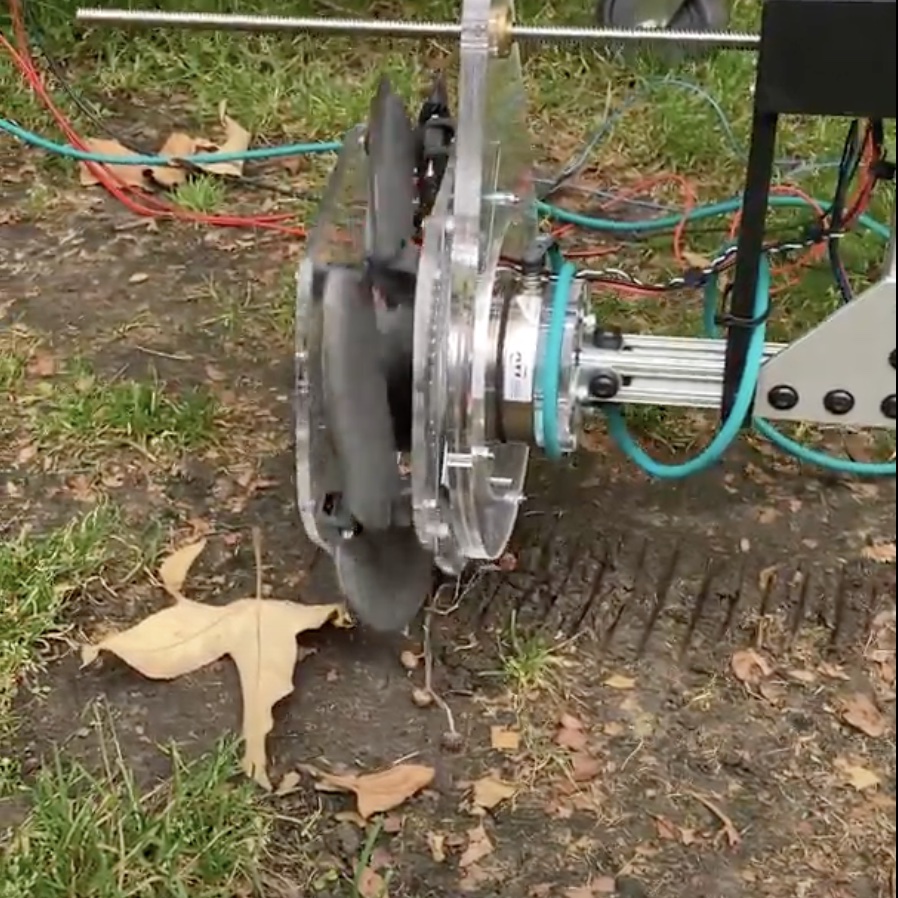}
    \includegraphics[width=0.48\linewidth, trim={5cm 2cm 0 1.5cm}, clip]{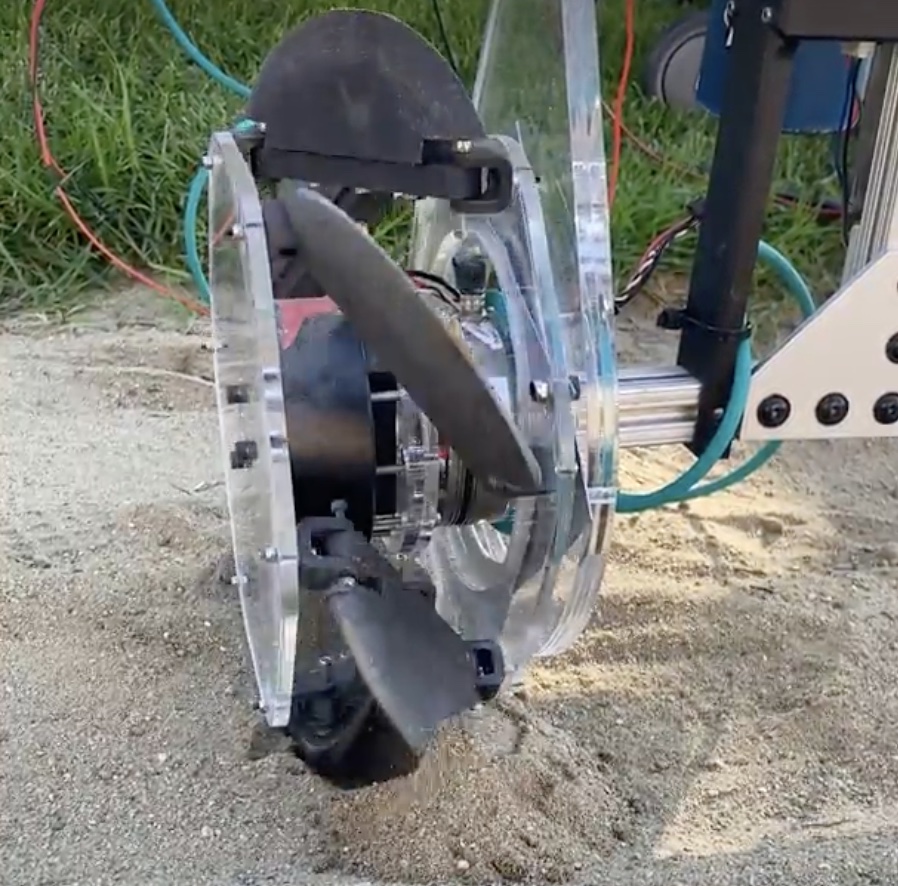}
    \caption{The image on the left shows NASU in mud, it leaves marks in the material as it provides a closer-to-solid surface to push off of when compressed, but is soft enough for the blades to pierce the media initially. The image on the right shows NASU in the sand, where the material is easy to pierce initially, but rather than compact, sand is instead dug out and pushed aside.}
    \label{sandvsmud}
\end{figure}

Meanwhile, NASU was unable to produce significant propulsion in sand which is different from our previous results in screw-based locomotion \cite{icra2023}.
This is likely due to two reasons.
First, NASU lacks an internal shell that is found on conventional Archimedes screw propellers.
The internal shell plays an important role in constraining the volume of material that is sheared by the screw blades so that the majority of the material is pushed along the axis of the screw.
This effect is especially pronounced in softer media with lower shearing forces such as sand. Secondly, the NASU has shorter, straight blades as opposed to conventional helical blades, which would provide a more continuous shearing action.
Figure \ref{sandvsmud} depicts the difference in the effectiveness of NASU in a compact media such as mud vs. a loose granular media such as sand. 

We hypothesized that there is a different optimal angle of attack, concerning velocity or efficiency, for different media rather than the simple trade-off trend we found.
Previous research in terramechanics screw-based propulsion suggests that the physical properties of different media should influence the performance \cite{Nagaoka_2010}.
Our results show that the velocity and efficiency trends are consistent across all media for the range of angles we tested.
We believe that we found no environment-specific trends due to the specific range of angles we tested, differences in the design of NASU compared to conventional screw propellers, or there simply are no environment-specific trends for screw-based locomotion.

\section{Conclusion}

This work showcases the first mechanism to dynamically reconfigure the angle of attack of a screw propeller to optimize locomotion capability on the fly. We conduct experiments in a variety of real-world environments, demonstrating a trade-off exists between traveling velocity and locomotive efficiency in the range of NASU's configuration space.
Our intention with this mechanism is to enable future screw-based vehicles to control the angle of attack and make adjustments depending on their environment and use case.
For example, a higher angle of attack is more optimal when speed and mobility are a priority as anticipated from previous literature \cite{icra2023, cole1961inquiry, dugoff1967model}.
Alternatively, a key finding of our experiments is that a lower angle of attack is preferred if thrust force and efficiency are a priority, such as when towing a heavy load.
While NASU was not tested in water or fluid environments, it is known that larger angles of attack, in the specific case of Archimedes-type screw propellers, produce more thrust and velocity in fluids \cite{cole1961inquiry, Mayfield_2015, lim2023amphibious}, and we hypothesize that this trend will continue with NASU.


In the application of quad-screw designs \cite{lugo2017conceptual, USF} and snake-like screw-propelled robots \cite{arcsnake_icra, arcsnake_tro}, the screws are also used similarly to normal wheels (i.e. the screw-blades are fully slipping).
This is a particularly useful mode of locomotion when the surface is solid such as concrete.
We anticipate that NASU can also be used for this locomotion mode and believe larger angles of attack will be better for this wheeling motion as the ratio of lateral to axial force is higher.

In future work, the design may be improved by adding an internal shell as discussed in the previous section, and stacking multiple NASU units together to achieve a longer more continuous blade interaction.
We also intend to experiment with different blade geometries through the replaceable blade connection.
Another future work is to integrate this technology into ARCSnake \cite{arcsnake_tro, lim2023amphibious} or other screw-based robots to improve their versatility and adaptability.


\section{Acknowledgements}

The authors would like to thank Yaohui Chen, Mandy Cheung, Peter Gavrilov, Hoi Man (Kevin) Lam, Casey Price, Dimitrious Schreiber, Nikhil Uday Shinde, Anne-Marie Shui, and Mingwei Yeoh for their support of the project.

\balance
\bibliographystyle{ieeetr}
\bibliography{refs}

\begin{thebibliography}{10}

\bibitem{rossell_chapman_1962}
H.~E. Rossell and L.~B. Chapman, ``Principles of naval architecture,'' {\em Published by the Society of Naval Architects and Marine Engineers}, 1962.

\bibitem{wells_1841}
T.~Wells, ``Improvement in the manner of constructing and of propelling steamboats, denominated the buoyant spiral propeller,'' U.S. Patent 2400, Dec. 1841.

\bibitem{neumeyer1965marsh}
M.~Neumeyer and B.~Jones, ``The marsh screw amphibian,'' {\em Journal of Terramechanics}, vol.~2, no.~4, pp.~83--88, 1965.

\bibitem{fales1972riverine}
W.~Fales, D.~W. Amick, and B.~G. Schreiner, ``The riverine utility craft (ruc),'' {\em Journal of Terramechanics}, vol.~8, no.~3, pp.~23--38, 1972.

\bibitem{Freeberg_2010}
J.~Freeberg, {\em A study of omnidirectional quad-screw-drive configurations for all-terrain locomotion}.
\newblock PhD thesis, University of South Florida, Oct. 2010.

\bibitem{villacres2023literature}
J.~Villacr{\'e}s, M.~Barczyk, and M.~Lipsett, ``Literature review on archimedean screw propulsion for off-road vehicles,'' {\em Journal of Terramechanics}, vol.~108, pp.~47--57, 2023.

\bibitem{li_2008}
R.~Li, B.~Wu, K.~Di, A.~Angelova, R.~E. Arvidson, I.-C. Lee, M.~Maimone, L.~H. Matthies, L.~Richer, R.~Sullivan, and et~al., ``Characterization of traverse slippage experienced by spirit rover on husband hill at gusev crater,'' {\em Journal of Geophysical Research}, vol.~113, no.~E12, 2008.

\bibitem{Nagaoka_2010}
K.~Nagaoka, M.~Otsuki, T.~Kubota, and S.~Tanaka, ``Terramechanics-based propulsive characteristics of mobile robot driven by {{Archimedean}} screw mechanism on soft soil,'' in {\em 2010 {{IEEE}}/{{RSJ International Conference}} on {{Intelligent Robots}} and {{Systems}}}, pp.~4946--4951, Oct. 2010.

\bibitem{osinski_2015}
D.~Osi{\'n}ski and K.~Szykiedans, ``Small remotely operated screw-propelled vehicle,'' in {\em Progress in Automation, Robotics and Measuring Techniques: Volume 2 Robotics}, pp.~191--200, Springer, 2015.

\bibitem{USF}
J.~T. Freeberg, {\em A Study of Omnidirectional Quad-Screw-Drive Configurations for All-Terrain Locomotion}.
\newblock PhD thesis, University of South Florida, 2010.

\bibitem{lugo2017conceptual}
J.~H. Lugo, V.~Ramadoss, M.~Zoppi, and R.~Molfino, ``Conceptual design of tetrad-screw propelled omnidirectional all-terrain mobile robot,'' in {\em 2nd International Conference on Control and Robotics Engineering}, pp.~13--17, IEEE, 2017.

\bibitem{arcsnake_icra}
D.~A. Schreiber, F.~Richter, A.~Bilan, P.~V. Gavrilov, H.~M. Lam, C.~H. Price, K.~C. Carpenter, and M.~C. Yip, ``Arcsnake: an archimedes’ screw-propelled, reconfigurable serpentine robot for complex environments,'' in {\em 2020 IEEE International Conference on Robotics and Automation (ICRA)}, pp.~7029--7034, IEEE, 2020.

\bibitem{arcsnake_tro}
F.~Richter, P.~V. Gavrilov, H.~M. Lam, A.~Degani, and M.~C. Yip, ``Arcsnake: Reconfigurable snakelike robot with archimedean screw propulsion for multidomain mobility,'' {\em IEEE Transactions on Robotics}, vol.~38, no.~2, pp.~797--809, 2021.

\bibitem{lim2023amphibious}
J.~Lim, {\em Amphibious Locomotion with a Screw-propelled Snake-like Robot}.
\newblock MSc Thesis, University of California, San Diego, 2023.

\bibitem{thakker2023eels}
R.~Thakker, M.~Paton, M.~P. Strub, M.~Swan, G.~Daddi, R.~Royce, P.~Tosi, M.~Gildner, T.~Vaquero, M.~Veismann, {\em et~al.}, ``Eels: Towards autonomous mobility in extreme terrain with a versatile snake robot with resilience to exteroception failures,'' in {\em 2023 IEEE/RSJ International Conference on Intelligent Robots and Systems (IROS)}, pp.~9886--9893, IEEE, 2023.

\bibitem{gildner2024boldly}
M.~Gildner, N.~Georgiev, E.~Ambrose, T.~Pailevanian, A.~Archanian, H.~Melikyan, D.~Loret~de Mola~Lemus, M.~Paton, R.~Thakker, and M.~Ono, ``To boldly go where no robots have gone before--part 2: The versatile mobility of the eels robot for robustly exploring unknown environments,'' in {\em American Institute of Aeronautics and Astronautics SciTech Forum}, p.~1965, 2024.

\bibitem{carpenter2021exobiology}
K.~Carpenter, A.~Thoesen, D.~Mick, J.~Martia, M.~Cable, K.~Mitchell, S.~Hovsepian, J.~Jasper, N.~Georgiev, R.~Thakker, A.~Kourchians, B.~Wilcox, M.~Yip, and H.~Marvi, {\em Exobiology Extant Life Surveyor (EELS)}, pp.~328--338.
\newblock ASCE Library, 2021.

\bibitem{cole1961inquiry}
A.~M. Group and B.~Cole, ``Inquiry into amphibious screw traction,'' {\em Proceedings of the Institution of Mechanical Engineers}, vol.~175, no.~1, pp.~919--940, 1961.

\bibitem{dugoff1967model}
H.~Dugoff and I.~R. Ehlich, ``Model tests of bouyant screw rotor configurations,'' {\em Journal of Terramechanics}, vol.~4, no.~3, pp.~9--22, 1967.

\bibitem{novelino2020untethered}
L.~S. Novelino, Q.~Ze, S.~Wu, G.~H. Paulino, and R.~Zhao, ``Untethered control of functional origami microrobots with distributed actuation,'' {\em Proceedings of the National Academy of Sciences}, vol.~117, no.~39, pp.~24096--24101, 2020.

\bibitem{Miyazawa2022}
Y.~Miyazawa, C.-W. Chen, R.~Chaunsali, T.~S. Gormley, G.~Yin, G.~Theocharis, and J.~Yang, ``Topological state transfer in kresling origami,'' {\em Comunication Materials}, 2022.

\bibitem{ze2022soft}
Q.~Ze, S.~Wu, J.~Nishikaea, J.~Dai, Y.~Sun, S.~Leanza, C.~Zemelka, L.~S. Novelino, G.~H. Paulino, and R.~R. Zhao, ``Soft robotic origami crawler,'' {\em Science Advances}, vol.~8, no.~13, p.~eabm7834, 2022.

\bibitem{PhysRevE.101.063003}
N.~Kidambi and K.~W. Wang, ``Dynamics of kresling origami deployment,'' {\em Physical Review E}, vol.~101, p.~063003, Jun 2020.

\bibitem{icra2023}
J.~Lim, C.~Joyce, E.~Peiros, M.~Yeoh, P.~V. Gavrilov, S.~G. Wickenhiser, D.~A. Schreiber, F.~Richter, and M.~C. Yip, ``Mobility analysis of screw-based locomotion and propulsion in various media.,'' {\em 2023 IEEE International Conference on Robotics and Automation (ICRA)}, 2023.

\bibitem{marvi_2018}
A.~Thoesen, S.~Ramirez, and H.~Marvi, ``Screw-powered propulsion in granular media: An experimental and computational study,'' {\em 2018 IEEE International Conference on Robotics and Automation (ICRA)}, 2018.

\bibitem{marvi_2019}
A.~Thoesen, S.~Ramirez, and H.~Marvi, ``Screw‐generated forces in granular media: Experimental, computational, and analytical comparison,'' {\em American Institute of Chemical Engineers Journal}, vol.~65, no.~3, p.~894–903, 2019.

\bibitem{Mayfield_2015}
W.~H. Mayfield, {\em Development of a {{Novel Amphibious Locomotion System}} for Use in {{Intra-Luminal Surgical Procedures}}}.
\newblock PhD thesis, University of Leeds, Aug. 2015.

\bibitem{seo_2021}
C.~Seo, K.~Lee, D.~Son, and T.~Seo, ``Robust design of a screw-based crawling robot on a granular surface,'' {\em IEEE Access}, vol.~9, p.~103988–103995, 2021.

\bibitem{doi:10.1177/1045389X11414084}
S.~Barbarino, O.~Bilgen, R.~M. Ajaj, M.~I. Friswell, and D.~J. Inman, ``A review of morphing aircraft,'' {\em Journal of Intelligent Material Systems and Structures}, vol.~22, no.~9, pp.~823--877, 2011.

\bibitem{doi:10.1098/rsif.2017.0240}
D.~D. Chin, L.~Y. Matloff, A.~K. Stowers, E.~R. Tucci, and D.~Lentink, ``Inspiration for wing design: how forelimb specialization enables active flight in modern vertebrates,'' {\em Journal of The Royal Society Interface}, vol.~14, no.~131, p.~20170240, 2017.

\bibitem{Wichita}
J.~Dorfling, {\em Feasibilty of Morphing Aircraft Propeller Blades}.
\newblock PhD thesis, Wichita State University, 2008.

\bibitem{10.1115/1.4054249}
K.~Ye and J.~Ji, ``{A Novel Morphing Propeller System Inspired by Origami-Based Structure},'' {\em Journal of Mechanisms and Robotics}, vol.~15, p.~011006, 04 2022.

\bibitem{CHEN2017746}
F.~Chen, L.~Liu, X.~Lan, Q.~Li, J.~Leng, and Y.~Liu, ``The study on the morphing composite propeller for marine vehicle. part i: Design and numerical analysis,'' {\em Composite Structures}, vol.~168, pp.~746--757, 2017.

\bibitem{nagatani2007development}
K.~Nagatani, M.~Kuze, and K.~Yoshida, ``Development of transformable mobile robot with mechanism of variable wheel diameter,'' {\em Journal of Robotics and Mechatronics}, vol.~19, no.~3, pp.~252--257, 2007.

\bibitem{crawling}
F.~Chen, L.~Liu, X.~Lan, Q.~Li, L.~Jinsong, and Y.~Liu, ``The study on the morphing composite propeller for marine vehicle. part i: Design and numerical analysis,'' {\em Composite Structures}, vol.~168, 02 2017.

\end{thebibliography}
\balance

\end{document}